\documentclass[journal]{IEEEtran}
\usepackage[usenames]{color}
\usepackage{hyperref}
\usepackage{jumoline}
\usepackage{ulem}
\usepackage{proofread} 
\noproofreadmark
\usepackage{tipa}

\usepackage{cite}
\let\em\relax
\DeclareTextFontCommand{\em}{\it}
\let\emph\relax
\DeclareTextFontCommand{\emph}{\it}
\usepackage{graphicx}
\usepackage{bm} 
\usepackage{mathptmx} 
\usepackage{times} 
\usepackage{amsmath} 
\usepackage{amssymb}  
\usepackage{algorithm}
\usepackage{algorithmic}

\title{Double Articulation Analyzer with Prosody for Unsupervised Word and Phoneme Discovery}
\author{
Yasuaki~Okuda,
Ryo~Ozaki,
and~Tadahiro~Taniguchi
\thanks{Y. Okuda and R. Ozaki are with the Graduate School of Information Science and Engineering, Ritsumeikan University, 1-1-1 Noji Higashi, Kusatsu, Shiga 525-8577, Japan {\tt\small  okuda.yasuaki@em.ci.ritsumei.ac.jp, ryo.ozaki@em.ci.ritsumei.ac.jp}}
\thanks{T. Taniguchi is with College of Information Science and Engineering, Ritsumeikan University, 1-1-1 Noji Higashi, Kusatsu, Shiga 525-8577, Japan {\tt\small  taniguchi@em.ci.ritsumei.ac.jp}}
}

\begin{document}
\maketitle

\begin{abstract}
Word and phoneme discovery are important tasks in language development for human infants. Infants acquire words and phonemes from unsegmented speech signals using segmentation cues, such as distributional, prosodic, and co-occurrence cues.
Many pre-existing computational models that represent the process tend to focus on distributional or prosodic cues.
This paper proposes a nonparametric Bayesian probabilistic generative model called the prosodic hierarchical Dirichlet process-hidden language model (Prosodic HDP-HLM) for simultaneous phoneme and word discovery from continuous speech signals. Prosodic HDP-HLM, an extension of HDP-HLM, considers both prosodic and distributional cues within a single integrative generative model. We propose a prosodic double articulation analyzer (Prosodic DAA) by deriving an inference procedure for Prosodic HDP-HLM.  
We conducted three experiments on different types of datasets, i.e., Japanese vowel sequence, utterances for teaching object names and features, and utterances following Zipf's law and demonstrate the validity of the proposed method.
The results show that the Prosodic DAA successfully uses prosodic cues and outperforms a method that solely uses distributional cues. In contrast, the phoneme discovery performance did not improve.
We also show that prosodic cues contributed to word discovery performance more when the word frequency was distributed more naturally, i.e., following Zipf's law.
The main contributions of this study are as follows: 1) We develop a probabilistic generative model for time series data including prosody that potentially has a double articulation structure; 2) We propose the Prosodic DAA by deriving the inference procedure for Prosodic HDP-HLM and show that Prosodic DAA can discover words directly from continuous human speech signals using statistical information and prosodic information in an unsupervised manner; 3) We show that prosodic cues contribute to word segmentation more in naturally distributed case words, i.e., they follow Zipf's law.
\end{abstract}


\IEEEpeerreviewmaketitle


\begingroup
\renewcommand{\thefootnote}{}
\footnotetext{This research was partially supported by Grants-in-Aid for Scientific Research on Innovative Areas (16H06569) funded by the Ministry of Education, Culture, Sports, Science, and Technology, Japan. }
\endgroup

\section{Introduction} \label{sec:Introduction}

\IEEEPARstart{S}PEECH signal segmentation problems that identify word and phoneme boundaries from continuous speech using segmentation cues, e.g., distributional cues and prosodic cues, are essential for human infant language acquisition.
This task is easy if speech signals are always given as a single word. However, many infant-directed speeches are known to consist of multiple words~\cite{aslin1996models,thiessen2005infant}. Nevertheless, human infants can discover words and phonemes from raw continuous speech signals. This word discovery from continuous speech signals is a difficult task because infants cannot use any information that explicitly identifies the boundaries of words but only uses cues contained in continuous speech signals, i.e., unsupervised learning. In addition, phoneme discovery needs to be performed using speech signals in an unsupervised manner as well.

Human infants can exploit numerous cues to discover words from continuous speech signals in the language acquisition process~\cite{pelucchi2009statistical}. These cues are 1) distributional, 2) prosodic, 3) co-occurrence, and other cues. 1) The distributional cues represent the statistical relationships that one element of speech sound follows another. 2) The prosodic cues rely on acoustic information, such as silent pause, stressed syllables, rhythmic bias, and suprasegmental features, e.g., pitch~\cite{jusczyk1993infants,jusczyk1999beginnings,morgan1996rhythmic,choi2003young,mugitani2019use}. 3) Co-occurrence cues represent events and objects observed in accordance with the utterance of a word.
It has been reported that 8-month-old infants can discover words from fluent speech based solely on distributional cues~\cite{saffran1996statistical,johnson2001word}. It has also been reported that 7-month-old infants can discover words from fluent speech based on distributional cues rather than prosodic cues~\cite{thiessen2003cues}.
In contrast, it has also been reported that 2-month-old infants can perform word discovery using prosodic information~\cite{mandel1994does}. As a result, considering both distributional and prosodic cues is crucial for developing an unsupervised phoneme and word discovery model.

Several computational models for simultaneous unsupervised word and phoneme discovery using distributional cues have been developed~\cite{lee2015unsupervised,kamper2019truly,taniguchi2016nonparametric}. Taniguchi et al. proposed a probabilistic generative model for simultaneous word and phoneme discovery, called the hierarchical Dirichlet process hidden language model (HDP-HLM) and its inference algorithm~\cite{taniguchi2016nonparametric}. The HDP-HLM is a probabilistic generative model for time series data that potentially has a double articulation structure, i.e., hierarchically organized latent words and phonemes embedded in speech signals. An unsupervised learning method called the nonparametric Bayesian double articulation analyzer (NPB-DAA) was proposed based on HDP-HLM. The NPB-DAA can estimate the double articulation structure and discover words and phonemes that simultaneously are acquired in acoustic and language models. However, their performance is still limited.

Prosodic cues are also essential for word discovery~\cite{saffran1996word} and have been reported to be effective in machine learning methods for word segmentation~\cite{ludusan2015prosodic,ludusan2016role}. However, a probabilistic generative model that effectively integrates prosodic and distributional information has not yet been proposed. 
Therefore, in this study, we focus on the introduction of prosodic cues that are believed to help word discovery as supplemental cues. We develop an unsupervised machine learning method that can discover words and phonemes directly from continuous speech signals using distributional and prosodic cues by extending the NPB-DAA. Furthermore, we also propose an unsupervised learning method called prosodic double articulation analyzer (Prosodic DAA) and a probabilistic generative model called the prosodic hierarchical Dirichlet process hidden language model (Prosodic HDP-HLM) and its inference algorithm.

In what cases do prosodic cues especially contribute to word segmentation from the viewpoint of a statistical model of word discovery? This is an important question to ask. If the words are distributed evenly enough, distributional cues may be sufficient for word segmentation. However, it is widely known that word distribution follows Zipf's law ~\cite{zipf2016human}. 
Zipf's law is an empirical law found in many types of social, physical, and other scientific domains. The frequency of a word is inversely proportional to its rank in the frequency table of words. This means that there are so many words that are far less frequently observed than other frequently observed words. It is naturally considered that capturing distributional cues from such data is more difficult than capturing data in which every word is observed in an equally frequent manner. We hypothesized that prosodic cues contribute to word discovery performance more when words are distributed more naturally than artificially prepared datasets, i.e., in adherence to Zipf's law.

The main contributions of this paper are as follows: 1) We develop a probabilistic generative model for time series data including prosody that potentially has a double articulation structure and propose the Prosodic DAA by deriving the inference procedure for Prosodic HDP-HLM. 2) We show that the Prosodic DAA can discover words directly from continuous human speech signals using statistical information and prosodic information in an unsupervised manner. 3) We show that prosodic cues contribute to word segmentation more in case words are naturally distributed, i.e., they follow Zipf's law.

The remainder of the paper is structured as follows.
Section~\ref{sec:Background} describes the background of the proposed method. Section~\ref{sec:Prosodic_DAA} presents the prosodic HDP-HLM by extending HDP-HLM, describes the inference procedure of prosodic HDP-HLM, and proposes Prosodic DAA. Section~\ref{sec:Experiment 1},  Section~\ref{sec:Experiment 2}, and Section~\ref{sec:Experiment 3} evaluate the performance of the proposed method using Japanese vowel sequence utterances for teaching object names, features, and utterances following Zipf's law. Section~\ref{sec:Conclusion} concludes the paper.

\begin{figure*}[!ht]
  \begin{center}
    \includegraphics[width=160mm]{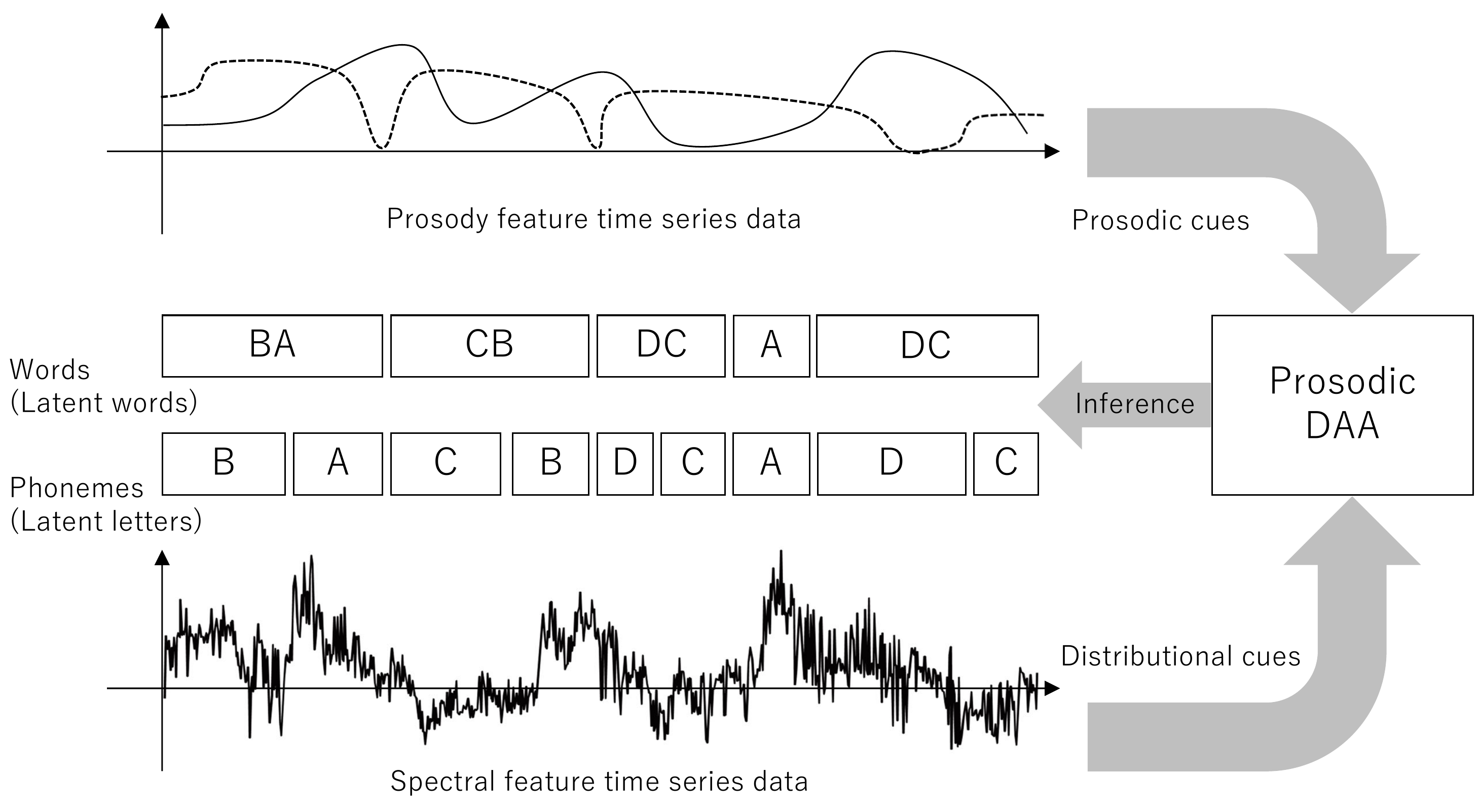}
    \caption{An illustrative overview of the proposed method}
    \label{fig:fig1}
  \end{center}
\end{figure*}

\section{Background} \label{sec:Background}
Human infants can use prosodic cues to discover words in the language acquisition process, as mentioned above. Prosodic cues have been shown to help word segmentation in language acquisition~\cite{christophe1996bootstrapping}. Based on this, Ludusan et al. extended the unsupervised word segmentation method~\cite{goldwater2009bayesian} to use prosodic cues and showed that prosodic cues help unsupervised word segmentation~\cite{ludusan2015prosodic,ludusan2016role}.
However, a computational model based on a probabilistic generative model, which makes use of distributional and prosodic cues jointly for simultaneous phoneme and word discovery, has not been proposed.

The unsupervised learning of an acoustic model, i.e., phoneme discovery, is a clustering task of feature vectors obtained from continuous speech signals. Mixture models, such as the Gaussian mixture models and hidden Markov models, have been used to categorize feature vectors of phonemes ~\cite{vallabha2007unsupervised,lake2009modeling,lee2012nonparametric,lee2013joint,siu2014unsupervised}. Phoneme acquisition is a complex categorization task in a feature space because of the overlap of the distribution of the feature vectors of each phoneme. The actual sound of a phoneme depends on its context. The importance of feedback information from segmented words in phoneme acquisition has been reported~\cite{feldman2013role}. Therefore, simultaneous word and phoneme discovery is essential. 

Statistical unsupervised simultaneous learning methods of acoustic and language models have been proposed~\cite{brandl2008self,walter2013hierarchical,lee2015unsupervised,kamper2019truly}. Word segmentation and phoneme categorization are mutually dependent, as pointed out in~\cite{feldman2013role}. Therefore, an integrated probabilistic generative model for unsupervised simultaneous learning of acoustic and language models is preferable. 
The unsupervised phoneme and word discovery from raw continuous speech signals can be regarded as an analysis of the double articulation of time series data. Double articulation is a hierarchical latent structure in which states, corresponding to words in language, have transitions among them in a stochastic manner at the higher level and those at the lower level, e.g., phonemes, have transitions in a deterministic manner inside the high-level state; for example, a word has a deterministic sequence of phonemes. Therefore, the phoneme and word discovery problem can be regarded as a double articulation analysis problem~\cite{taniguchi2011double,taniguchi2016nonparametric,Taniguchi2016SER}.
For double articulation analysis, Taniguchi et al. developed the NPB-DAA, which integrates the phoneme and word discovery processes into a single inference process of a unified generative model called HDP-HLM ~\cite{taniguchi2016nonparametric}\footnote{Another hierarchical Bayesian approach to acoustic signals involving acoustic and language models can be found in music analysis~\cite{ojima2018chord}.}. They showed that it could achieve phoneme and word discovery to some extent.
However, HDP-HLM only models distributional cues and does not use prosodic cues, such as silent pause and fundamental frequency. This paper proposes Prosodic HDP-HLM, which models distribution and prosodic cues related to word segments by extending the HDP-HLM.

The Zero Resource Speech Challenges aim to construct a system that learns an end-to-end spoken dialog system, using only the information available in the language acquisition process, have been organized~\cite{dunbar2017zero,renshaw2015comparison}. Probabilistic computational models that achieved unsupervised direct word discovery from continuous speech signals were proposed in the Zero Resource Speech Challenges. Kamper et al. proposed an unsupervised word segmentation system that segments and clusters speech data into a unit, such as a word~\cite{kamper2017segmental}. Recently, methods involving representation learning have also been developed~\cite{vanniekerk2020vectorquantized}. However, an integrative probabilistic generative model, especially based on Bayesian nonparametrics, involving prosodic and distributional cues, has not been proposed. 

In robotics, unsupervised word discovery methods have been studied to achieve online lexical acquisition and overcome the out-of-vocabulary problem~\cite{Araki2012,TomoakiNakamura2014,takeda2018word,takeda2017unsupervised,AkiraTaniguchi2018,Akira2020,taniguchi2016spatial,taniguchi2017online}. Several models use co-occurrence cues (e.g., object and place categories) to improve word discovery performance~\cite{TomoakiNakamura2014,taniguchi2017online,AkiraTaniguchi2018,Akira2020}. However, phoneme discovery and prosodic cues have rarely been considered.

Based on the above background, we propose a probabilistic generative model called the prosodic HDP-HLM for time series data, including prosody that potentially has a double articulation structure, containing both an acoustic model and language model. The latent double articulation structure of time series data can be analyzed in an unsupervised manner by assuming prosodic HDP-HLM as a generative model of observation data and inferring latent variables of the prosodic HDP-HLM. The unsupervised machine learning method for double articulation analysis is called Prosodic DAA. An overview of the proposed method is presented in Fig.~\ref{fig:fig1}).
The Prosodic DAA uses distributional cues and prosodic cues simultaneously in an explicit manner.

\section{Prosodic DAA} \label{sec:Prosodic_DAA}
\subsection{Generative Model: Prosodic HDP-HLM}\label{sec:Generative Model}
This section describes a probabilistic generative model, the prosodic HDP-HLM, by adding auxiliary observations corresponding to prosody to HDP-HLM, which is a nonparametric Bayesian probabilistic generative model for time series data that potentially has a double articulation structure.

A graphical model of the prosodic HDP-HLM is shown in Fig.~\ref{fig:phdphlm}). 
Notably, most of the generative processes are the same as HDP-HLM except for variables related to prosodic features $\{Y_t\}$.

The generative process of the prosodic HDP-HLM is described as follows:

\begin{figure*}[!t]
  \begin{center}
    \includegraphics[width=130mm]{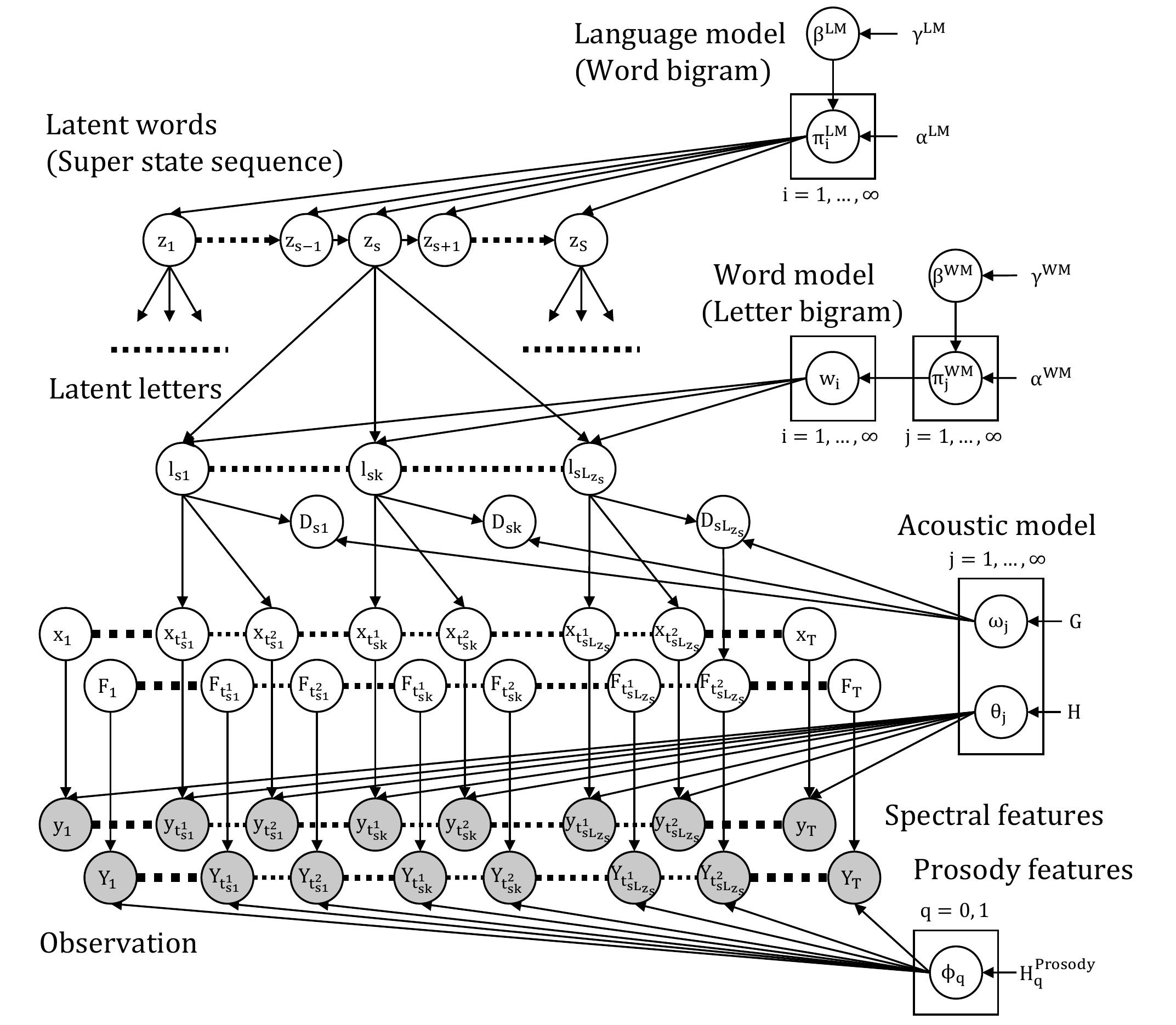}
    \caption{A graphical model for the proposed Prosodic HDP-HLM}
    \label{fig:phdphlm}
  \end{center}
\end{figure*} 
\begin{eqnarray}
  \beta^{\mathrm{LM}} &\sim& \rm GEM(\gamma^{\mathrm{LM}}) \\
  \pi_i^{\mathrm{LM}} &\sim& \rm DP(\alpha^{\mathrm{LM}},\beta^{\mathrm{LM}})\quad(i=1,\ldots) \\
  \beta^{\mathrm{WM}} &\sim& \rm GEM(\gamma^{\mathrm{WM}}) \\
  \pi_i^{\mathrm{WM}} &\sim& \rm DP(\alpha^{\mathrm{WM}},\beta^{\mathrm{WM}})\quad(i=1,\ldots) \\
  w_{i,k} &\sim& \pi_{w_{i,k-1}}^{\mathrm{WM}}\quad(i=1,\ldots)(k=1,\ldots,L_i) \\
  (\theta_j,\omega_j) &\sim& H \times G\quad(j=1,\ldots) \\
  \phi_{q} &\sim& H_{q}^{\rm Prosody}\quad(q=0,1) \\
  z_s &\sim& \pi_{z_{s-1}}^{\mathrm{LM}}\quad(s=1,\ldots,S)\\
  l_{sk} &=& w_{w_{z_sk}}\quad(s=1,\ldots,S)(k=1,\ldots,L_{z_s}) \\
  D_{sk} &\sim& g(\omega_{l_{sk}})\quad(s=1,\ldots,S)(k=1,\ldots,L_{z_s}) \\
  x_t &=& l_{sk}\quad(t=t^1_{sk},\ldots,t^2_{sk})\\
  t^1_{sk}&=&\sum_{s'<s}D_{s'}+\sum_{k'<k}D_{sk'}+1\nonumber\\
  t^2_{sk}&=&t^1_{sk}+D_{sk}-1\nonumber\\
  D_{s}^{\rm sum} &=& \sum_{s'\leq s}D_{s'} \\
  y_t &\sim& h(\theta_{x_t})\quad(t=1,\ldots,T)
\end{eqnarray}
\begin{eqnarray}
  F_t &=& \left\{\begin{array}{l}{0\quad(t=t_{s1}^1:D_{s}^{\rm sum}-1)}\\{1\quad(t=D_{s}^{\rm sum})}\end{array}\right.\\
  Y_t &\sim& h^{\rm Prosody}(\phi_{F_t})
\end{eqnarray}

where $\rm GEM$ and $\rm DP$ represent the stick-breaking and Dirichlet processes, respectively. $\rm LM$ represents the language model and $\rm WM$ represents the word model. The parameters $\gamma^{\rm WM}$ and $\alpha^{\rm WM}$ are hyperparameters of the word model, $\beta^{\rm WM}$ is a global transition probability that becomes the base measure of the transition probability distributions, and $\pi_j^{\rm WM}$ represents the transition probability from latent letter $j$ to the next latent letter. The parameters $\gamma^{\rm LM}$ and $\alpha^{\rm LM}$ are hyperparameters of the language model, $\beta^{\rm LM}$ is a global transition probability that becomes the base measure of the transition probability distributions, and $\pi_i^{\rm LM}$ represents the transition probability from latent word $i$ to the next latent word. The superscripts LM and WM indicate the language and word models, respectively.

The latent letter\footnote{In the terminology of double articulation analysis, the elements comprising a latent word are called latent letters. In phoneme and word discovery, a latent letter is considered an inferred index of a phoneme.} sequence of the $i$-th latent word $w_i$ is sampled from $\pi_{w_{i,k-1}}^{\rm WM}$. The duration distribution $g$ and observation distribution $h$ have parameters $\omega_j$ relating to the $j$-th latent letter and $\theta_j$ generated from the base measures $G$ and $H$. In addition, the prosodic observation distribution $h^{\rm Prosody}$ has parameters $\phi_q$ generated from the base measures $H^{Prosody}_q$. The variable $z_s$ is the $s$-th latent word in the latent word sequence and corresponds to the superstate in the hierarchical Dirichlet process hidden semiMarkov model (HDP-HSMM)~\cite{johnson2013bayesian}. The duration time $D_s$ is the frame duration of the $s$th latent word $z_s$. The latent letter $l_{sk} = w_{{z_s}k}$ corresponds to the $k$th latent letter of the $s$th latent word. The duration time $D_{sk}$ is the frame duration of the latent letter $l_{sk}$. The duration time $D_{s}^{\rm sum}$ is the frame duration from $t=1$ to the end point of the word $z_s$. The variables $x_t$ and $y_t$ indicate the hidden state and observation data at time $t$, respectively. 
In word and phoneme discovery, we assume that $y_t$ represents the spectral feature representation, for example, the mel-frequency cepstral coefficient (MFCC). 
The time frames $t^1_{sk}$ and $t^2_{sk}$ are the start and end points of the segment corresponding to $l_{sk}$, respectively.

The duration time $D_{sk}$ of the $k$-th latent letter $l_{sk}$ of the $s$-th latent word $z_s$ in the word sequence is drawn from the duration distribution $g(\omega_{l_{sk}})$. The duration of the latent word $z_s$ is $D_s = \sum_{k=1}^{L_{z_s}}D_{sk}$, where assuming $g$ is a Poisson distribution, the duration distribution of a latent word $z_s$ also follows a Poisson distribution because of the reproductive property of the Poisson distribution. In this case, the Poisson distribution parameter of the duration of the latent word is $\sum_{k=1}^{L_{z_s}}\omega_{l_{sk}}$.

In addition to the variables described above, which are also in the HDP-HLM, the prosodic HDP-HLM has additional prosody-related variables. 
The variables $Y_t$ and $F_t$ are prosodic observation data at time $t$ and indicate that a new word begins at $t+1$ when $F_t = 1$.
In this case, $F_t = 1$ when $t = D_{s}^{\rm sum}$. The parameter $q$ is the variable relating to the value of 0 or 1 in indicator $F_t$.
We assume that $Y_t$ is a prosodic feature observed in accordance with the word boundaries, i.e., $F_t = 1$.

\subsection{Inference procedure} \label{sec:Inference Algorithm}
The approximated blocked Gibbs sampler for the prosodic HDP-HLM can be derived in the same way as the approximated blocked Gibbs sampler for the HDP-HLM. The inference procedure of HDP-HLM, called NPB-DAA, can estimate the double articulation structure from time series data. The Prosodic HDP-HLM can find latent words and letters from time series data, including prosody, in an unsupervised manner, by inferring the latent local and global parameters of prosodic HDP-HLM.

In the HDP-HLM, we adopted the backward filtering forward-sampling procedure, which is the inference method of HDP-HSMM adapted to HDP-HLM. By extending the backward filtering forward-sampling procedure of HDP-HLM, we can obtain an inference procedure for prosodic HDP-HLM. The calculation of the backward messages of the latent word $z_s = i$ in prosodic HDP-HLM is as follows:

\begin{eqnarray}
 \beta_t(i)&:=&p(y_{t+1:T},Y_{t+1:T}|z_{s(t)}=i,F_t=1) \nonumber\\
 &=&\sum_{j}\beta_t^*(j)p(z_{s(t+1)}=j|z_t=i) \\
 \beta_t^*(i)&:=&p(y_{t+1:T},Y_{t+1:T}|z_{s(t+1)}=i,F_t=1) \nonumber\\
 &=&\sum_{d=1}^{T-t}\beta_{t+d}(i)p(D_{t+1}=d|z_{s(t+1)}=i) \nonumber\\\label{p(y)}
 &&\times p(y_{t+1:t+d},Y_{t+1:t+d}|i,d)\\
 \beta_T(i)&:=&1
\end{eqnarray}

where $z_{s(t)}$ represents the latent word $z_s$ at time $t$ and $D_{t+1}$ represents the duration of the latent word beginning at time $t+1$. The probability $\beta_t(i)$ is obtained by marginalizing all latent words $j$ at time $t+1$. The probability $\beta_t^*(i)$ is the probability that the latent word $i$ begins at time $t+1$. This probability $\beta_t^*(i)$ is obtained by marginalizing all duration frames $d$. The probability $p(y_{t+1:t+d},Y_{t+1:t+d}|i,d)$ in (\ref{p(y)}) shows the probability that observations $y_{t+1:t+d}$ and prosodic observations $Y_{t+1:t+d}$ are generated by the latent word $i$. The likelihood of the latent word $p(y_{t+1:t+d},Y_{t+1:t+d}|i,d)$ is as follows:

\begin{multline}
\label{dynamic}
 p(y_{t+1:t+d},Y_{t+1:t+d}|i,d)\\=\left(
\sum_{r\in R^{(L_i,d)}}\prod_{k=1}^{L_i}p(r_k|l_k) 
 \prod_{m=1}^{r_{k}}p(y_{t+m+\sum_{k^{\prime}=1}^{k-1}r_{k^{\prime}}}
 |l_{k})
 \right)\\
 \times \left(P(Y_{t+d}|F_{t+d}=1) \prod_{t'=1}^{d-1}P(Y_{t+t'} | F_{t+t'}=0) \right)
\end{multline}
\begin{eqnarray}
 R^{(L_i,d)}=\{r\in\{1,2,\ldots\}^{L_i}|\sum_{k=1}^{L_i}r_k=d\}
\end{eqnarray}
where the variable $R^{(L_i,d)}$ is a set of $L_i$-dimensional natural number vectors whose element summation  is $d$. The value of (\ref{dynamic}) can be calculated efficiently using dynamic programming. The forward message $\alpha_{t}(k)$ can be recursively calculated as follows:

\begin{eqnarray}
 \alpha_{t}(k)&=&\sum_{d^{\prime}=1}^{t-k+1}\alpha_{t-d^{\prime}}(k-1)p(d^{\prime}|l_{k})\nonumber\\
 &&\times\prod_{t^{\prime}=1}^{d^{\prime}}p(y_{t-t^{\prime}+1},Y_{t-t^{\prime}+1}|l_{k}) \\
 \alpha_{0}(0)&=&1
\end{eqnarray}

where the forward message $\alpha_{t}(k)$ is defined as the probability that the $k$-th latent letter in the latent word $w_i$ transitions to the next latent letter at time $t$. As a result, $\beta_t(i)$ and $\beta_t^*(i)$ can be calculated. The backward filtering forward-sampling procedure allows the blocked Gibbs sampler to directly sample latent words from observation data without explicitly sampling latent letters in prosodic HDP-HLM, similar to HDP-HLM. In the forward-sampling procedure, the latent word $z_{s(t+1)}$ and the duration $D_{s(t+1)}$ of the latent word $z_{s(t+1)}$ are sampled iteratively using backward messages as follows:

\begin{multline}
 p(z_s=i|y_{1:T},Y_{1:T},z_{s-1}=j,F_{D_{1:s}^{\mathrm{sum}}}=1)=\\
 p(i|j)\beta_{D_{1:s}^{\mathrm{sum}}}(i)p(y_{D_{1:s}^{\mathrm{sum}}},Y_{D_{1:s}^{\mathrm{sum}}}|i)
\end{multline}
\begin{multline}
 p(D_{s}=d|y_{1:T},Y_{1:T},z_{s}=i,F_{D_{1:s}^{\mathrm{sum}}}=1)=\\
 p(y_{D_{1:s}^{\mathrm{sum}}+1:D_{1:s}^{\mathrm{sum}}+d},Y_{D_{1:s}^{\mathrm{sum}}+1:D_{1:s}^{\mathrm{sum}}+d}|d, i,F_{D_{1:s}^{\mathrm{sum}}}=1)\\
 \times p(d)\frac{\beta_{D_{1:s}^{\mathrm{sum}}+d}(i)}{\beta_{D_{1:s}^{\mathrm{sum}}}^{*}(i)}
\end{multline}

where $D_{1:s}^{\mathrm{sum}} = \sum_{s'<s}D_{s'}$. From the calculation formula shown above, the latent word $z_{s(t+1)}$ and the duration $D_{s(t+1)}$ of the latent word $z_{s(t+1)}$ can be sampled using $\beta_t(i)$ and $\beta_t^*(i)$.

Once the latent words and their duration are sampled, the other parameters, e.g., model parameters and a latent letter sequence for each latent word, can be sampled in exactly the same way as the original HDP-HLM~\cite{taniguchi2016nonparametric}. For more details, please refer to the original paper~\cite{taniguchi2016nonparametric}.

\subsection{Prosodic DAA and prosody features}

 The inference procedure of Prosodic HDP-HLM allows the estimation of the double articulation structure from time series data. Therefore, we call the unsupervised machine learning method based on Prosodic HDP-HLM Prosodic DAA, in the same way as the unsupervised learning method that is based on the original HDP-HLM is called NPB-DAA\footnote{The source code of Prosodic DAA is available at: \\\indent \url{https://github.com/EmergentSystemLabStudent/Prosodic-DAA}}.

Generally, Prosodic DAA does not specify a feature extraction method for prosody features. Any prosody features that are informative for word boundaries can be used. In the experiment described later in this study, we use the fundamental frequency $\rm F_0$, and silent pauses are used as prosody observations. We focus on these two prosodic cues because they are likely universal cues for word discovery\cite{vaissiere2008}. The second-order differential of the fundamental frequency $\rm F_0$ and the duration of silent pause extracted from audio, instead of removing them, are given as prosodic feature observations $Y_t := (Y_{1,t}, Y_{2,t})$, respectively.
Further details of the feature extraction are described in the experimental section.

However, if another prosody feature co-occurring with a word boundary is prepared, such additional prosody features can be easily introduced into Prosodic DAA without any extension of the model. 

\section{Experiment 1: Continuous Japanese Vowel Speech Signal} \label{sec:Experiment 1}
In the first experiment, we evaluated Prosodic DAA using Japanese vowel speech signals to verify the applicability of the proposed method to actual human continuous speech signals.

The speech utterances in the dataset do not have rich prosody features and are monotonous. In addition, the word distributions are artificially designed and the distributional cues are relatively easy to find.  This dataset was used to examine whether the method could find words and phonemes using distributional cues. In this experiment, we evaluated whether the Prosodic DAA can perform word and phoneme discovery in the same way as NPB-DAA on this dataset.

In this experiment, we compared the proposed method Prosodic DAA and NPB-DAA~\cite{taniguchi2016nonparametric}, i.e., statistical word and phoneme discovery with and without prosodic cues.

\subsection{Conditions}

\begin{table*}[t]
\begin{center}
\caption{ARI for Estimated Latent Letters and Words in Experiment 1}
\label{tab:experiment1}
\begin{tabular}{lcccc}\hline
\textbf{Method} & \textbf{Phoneme ARI} & \textbf{Word ARI} & \textbf{Trained Acoustic Model} & \textbf{True Word Dictionary} \\ \hline
Prosodic DAA (fundamental frequency $\rm F_0$ + silent pause) &0.508$\pm$0.021 & \underline{\bf0.759$\pm$0.034} \\
Prosodic DAA (silent pause) & 0.519$\pm$0.034 & \bf0.722$\pm$0.059 \\
Prosodic DAA (fundamental frequency $\rm F_0$) & \underline{\bf0.521$\pm$0.029} & 0.718$\pm$0.096 \\
NPB-DAA & \bf0.511$\pm$0.030 & 0.667$\pm$0.093 \\ \hline
Julius GMM & 0.219 & 0.071 & \checkmark \\
Julius DNN & 0.120 & 0.160 & \checkmark \\ \hline
Julius GMM with true word dictionary & 0.365 & 0.340 & \checkmark & \checkmark \\
Julius DNN with true word dictionary & 0.324 & 0.565 & \checkmark & \checkmark \\ \hline
\end{tabular} 
\end{center}
\end{table*}

We used the same dataset\footnote{Japanese vowel native speech dataset
 :\\\indent \url{https://github.com/EmergentSystemLabStudent/aioi_dataset}} as in~\cite{taniguchi2016nonparametric,taniguchi2016double}.
 The data consisted of 60 audio files; a native female Japanese speaker read 30 artificial sentences aloud twice at a natural speed and recorded it. The sentences comprised five words \{aioi, aue, ao, ie, uo\}, which consisted of five Japanese vowels \{a, i, u, e, o\} representing {\{}{\textipa{\"a, i, W\super B, \|`e, \|`o}}{\}} in phonetic symbols respectively. By combining the 5 words, the 30 sentences include 25 two-word sentences, e.g., ``aioi aioi,'' ``aue ie,'' and ``uo ao,'' and five three-word sentences i.e., ``aioi uo ie,'' ``aue ao ie,'' ``ao ie ao,'' ``ie ie uo,'' and ``uo aue ie,'' were prepared. The set of two-word sentences consisted of all possible word pairs.

The data were encoded into 12-dimensional MFCC time-series data as observation data for spectral features. The frame size of MFCC was set to 25 ms, and the frame shift of MFCC was set to 10 ms. We used DSAE as an adaptive feature extractor in the same way as~\cite{taniguchi2016double} and extracted 3-dimensional data as observation data and the DSAE parameters $\alpha = 0.003$, $\beta = 0.7$, and $\eta = 0.5$. For more details, please refer to the original paper on NPB-DAA with DSAE~\cite{taniguchi2016double}.

The prosody features were extracted as follows. 
The second-order differential of the fundamental frequency $\rm F_0$ $(Y_{1,t})$ and duration of silent pause $(Y_{2,t})$ are extracted and used as time series data for prosody feature observations. Robust Epoch and Pitch EstimatoR (REAPER\footnote{REAPER : \url{https://github.com/google/REAPER}}) were used to extract the fundamental frequency $\rm F_0$ and the parameters of frame size and minimum and maximum $\rm F_0$ were set to 0.01, 40.0, and 300.0, respectively. A section where the volume below the threshold is continuous for a certain period is defined as a silent pause. Notably, the silent pauses were removed from the audio data. When the duration of the silent pause after time frame $t$ was detected and extracted, the duration $d^{\text{sil}}$ was set to $Y_{1,t} = d^{\text{sil}}$ representing the silent pause from the current frame to the next frame. The threshold of maximum volume and minimum period of silent pause were set to -8 dB and 0.01 s, respectively.

The hyperparameters for HDP-HLM and Prosodic HDP-HLM were set to $\gamma^{\rm LM} = 10.0$, $\alpha^{\rm LM} = 10.0$, $\gamma^{\rm WM} = 10.0$, and $\alpha^{\rm WM} = 10.0$. The hyperparameters of the duration distribution were set to $\alpha_0=200$ and $\beta_0=10$. The hyperparameters of the observation distribution were set to $\mu_0=0$, $\Sigma_0 = I$, $\kappa_0=0,01$, and $\nu_0=(\text{dimension}+2)$. The hyperparameters of the prosodic observation distribution were set to $\mu_0=0$, $\Sigma_0 = I$, $\kappa_0=100$, and $\nu_0=(\text{dimension}+2)$ for $F_t=0$, and $\mu_0=1$, $\Sigma_0 = I$, $\kappa_0=2$, and $\nu_0=(\text{dimension}+2)$ for $F_t=1$. In addition, we set the maximum number of letters and words and the maximum frame duration of words to 10 and 90 for weak-limit approximation. With different random number seeds, 20 trials with a Gibbs sampling of 100 iterations were performed.

In this experiment, we employed an open-source large vocabulary continuous speech recognition engine, Julius\footnote{Open-Source Large Vocabulary Continuous Speech Recognition\\\indent Engine Julius : \url{https://github.com/julius-speech/julius}}\cite{lee2001julius} as a baseline method for comparison with the proposed Prosodic DAA. The acoustic model of Julius was trained using a large speech dataset in a supervised manner. For the experiment, we used a GMM-based triphone model and a DNN-based triphone model.

We prepared two different groups of conditions for Julius. The first group used Julius because it is a generic speech recognition system. In this group, Julius used a generic word dictionary. The second group used the true word dictionary of the dataset. Therefore, in this group, Julius used the true word list and true phoneme list for continuous speech recognition.

\subsection{Results}

The phoneme and word discovery task, i.e., double articulation analysis, can be regarded as an unsupervised clustering task. We evaluated the experimental results using the adjusted rand index (ARI), which quantifies the performance of a clustering task. ARI becomes 1 when the clustering result matches the ground truth and becomes zero when the data are clustered randomly. We provided phonemes, i.e., latent letters and word ground truth labels, to all datasets and evaluated the experimental results.

In Table ~\ref{tab:experiment1}, phoneme ARI (i.e., the average ARI for latent letters) and word ARI (i.e., that for latent words) estimated by  NPB-DAA and proposed Prosodic DAA using only silent pause, only $\rm F_0$ and both $\rm F_0$ and silent pauses are shown. The ARI for estimated latent letters and words shows how accurately each method estimated latent letters and words, which correspond to phonemes and words in speech signals. A higher ARI indicates a more accurate estimation of latent variables. The experimental results showed an average ARI of 20 trials.

In Table ~\ref{tab:experiment1}, the ARI of latent letters, i.e., phonemes and latent words of two different groups of conditions for Julius; NPB-DAA; and proposed Prosodic DAA using only silent pause, only $\rm F_0$, and both $\rm F_0$ and silent pauses are shown. The ARI for estimated latent letters and words shows how accurately each method estimated latent letters and words, which correspond to phonemes and words in speech signals. A higher ARI indicates a more accurate estimation of latent variables. The experimental results of the DAAs show an average ARI of 20 trials.


The results show that the proposed Prosodic DAAs in all conditions outperformed NPB-DAA at word ARI. In contrast, there is almost no difference between Prosodic DAAs in all conditions and NPB-DAA at phoneme ARI.
Comparing Prosodic DAAs in all conditions with NPB-DAA and calculating the t-test at $p = 0.05$, Prosodic DAA using both $\rm F_0$ and silent pause, and Prosodic DAA using only silent pauses were statistically significantly different from NPB-DAA at word ARI ($p=3.6\times 10^{-4}$ and $p=3.4\times 10^{-2}$, respectively).
There were no statistically significant differences in all combinations of DAAs at phoneme ARI.

The statistically significant differences in word ARI between Prosodic DAA using only silent pause and NPB-DAA showed that the word segmentation performance improved when using prosodic cues.
Moreover, the statistically significant differences in word ARI between Prosodic DAA using both $\rm F_0$ and silent pause and Prosodic DAA using only silent pause showed that the word segmentation performance improved by using both $\rm F_0$ and silent pause instead of only silent pause.
The results of Julius in all conditions were lower than the results of DAAs. This is likely because the acoustic and language models assumed in the dataset in Experiment 1 and that in Julius are very different.
These results show that the Prosodic DAA improved the word discovery performance of NPB-DAA even when the target data had moderate prosody features.

\section{Experiment 2: Japanese Continuous Speech Signal including Prosody} \label{sec:Experiment 2}
In the second experiment, we evaluated our proposed method using naturally spoken Japanese speech signals, which contain consonants, ordinal Japanese vocabularies, and richer prosody features than the speech signals used in Experiment 1. 
This experiment was conducted to verify the applicability of the proposed method to actual human continuous Japanese utterances.

\subsection{Conditions}

\begin{table*}[ht]
\begin{center}
\caption{ARI for Estimated Latent Letters and Words in Experiment 2}
\label{tab:experiment2}
\begin{tabular}{lcccc}\hline
\textbf{Method} & \textbf{Phoneme ARI} & \textbf{Word ARI} & \textbf{Trained Acoustic Model} & \textbf{True Word Dictionary} \\ \hline
Prosodic DAA (fundamental frequency $\rm F_0$ + silent pause) &\bf0.369$\pm$0.017 & \underline{\bf0.717$\pm$0.030} \\
Prosodic DAA (silent pause) & \underline{\bf0.371$\pm$0.017} & \bf0.644$\pm$0.037 \\
Prosodic DAA (fundamental frequency $\rm F_0$) & 0.365$\pm$0.016 & 0.607$\pm$0.047 \\
NPB-DAA & 0.365$\pm$0.016 & 0.559$\pm$0.050 \\ \hline
Julius GMM & 0.575 & 0.557 & \checkmark \\
Julius DNN & 0.474 & 0.725 & \checkmark \\ \hline
Julius GMM with true word dictionary & 0.677 & 0.900 & \checkmark & \checkmark \\
Julius DNN with true word dictionary & 0.493 & 0.825 & \checkmark & \checkmark \\ \hline
\end{tabular} 
\end{center}
\end{table*}

We prepared the dataset of continuous Japanese utterances\footnote{Japanese native speech dataset
 :\\\indent \url{https://github.com/EmergentSystemLabStudent/object_teaching_dataset}}. The data consisted of 70 audio files; a native male Japanese speaker read 70 artificial sentences aloud once at a natural speed and recorded it. The data consisted of sentences that teach names and features of objects, for example, ``kore wa omocha'' in English ``This is a toy'' and ``yawarakai yo'' in English ``It's soft.'' The sentences comprised 26 words, consisting of 26 Japanese phonemes. 
We prepared phonemes, i.e., latent letters, and word ground truth labels, to all datasets and evaluated the relationship between the ground truth labels and estimated latent letters and words using the ARI. We used the automatic annotation tool provided by Julius GMM to prepare ground truth labels.

All the feature extraction methods and hyperparameters were used in the same way as in Experiment 1, except for the following.
The data were encoded as observation data into 36-dimensional MFCC, which is a concatenation of 12-dimensional MFCC, the differential of 12-dimensional MFCC, and second-order differential of 12-dimensional MFCC time series data.
We used DSAE as an adaptive feature extractor in the same way as~\cite{tada2017comparative} and extracted 9-dimensional data as observation data. For more details, please refer to the original paper on NPB-DAA with DSAE for natural speech signals~\cite{tada2017comparative}.
To extract prosody features, the threshold of the maximum volume and minimum period of silent pause was set to -10 dB and 0.03 s, respectively.

Regarding the hyperparameters for HDP-HLM and Prosodic HDP-HLM, the maximum number of letters and words and the maximum frame duration of words were set to 50 and 120 for weak-limit approximation.

In this experiment, we employed two different groups of conditions for Julius as a baseline method, similar to Experiment 1.

\subsection{Results}
In Table ~\ref{tab:experiment2}, the ARI of latent letters, i.e., phonemes, and latent words of two different groups of conditions for Julius, NPB-DAA, and proposed Prosodic DAA using only silent pause, only $\rm F_0$; and both $\rm F_0$ and silent pauses are shown. The ARI for estimated latent letters and words shows how accurately each method estimated latent letters and words, which correspond to phonemes and words in speech signals. A higher ARI indicates a more accurate estimation of latent variables. The experimental results of the DAAs show an average ARI of 20 trials.

The results show that the proposed Prosodic DAAs in all conditions outperformed NPB-DAA at word ARI. In contrast, there is almost no difference between Prosodic DAAs in all conditions and NPB-DAA at phoneme ARI.
Comparing Prosodic DAAs in all conditions with NPB-DAA and calculating the t-test at $p = 0.05$, there were statistically significant differences in all combinations of DAAs at word ARI ($p=2.4\times 10^{-13}$, $4.6\times 10^{-07}$, and $3.2\times and 10^{-4}$, for Prosodic DAA with both features, $\rm F_0$, and silent pause, respectively ). There were no statistically significant differences in any combination of DAAs at phoneme ARI.

The statistically significant differences in word ARI between Prosodic DAA using either $\rm F_0$ or silent pause and NPB-DAA showed that the word segmentation performance improved when using prosodic cues.
The statistically significant differences in word ARI between Prosodic DAA using both $\rm F_0$ and silent pause and Prosodic DAA using either $\rm F_0$ or silent pause showed that the word segmentation performance improved by using both $\rm F_0$ and silent pause instead of either $\rm F_0$ or silent pause.
In addition, the statistically significant difference in word ARI between Prosodic DAA using only silent pause and Prosodic DAA using only $\rm F_0$ showed that the word segmentation performance was improved by using silent pause instead of $\rm F_0$.
Notably, the results of several ARIs of Julius GMM outperform Julius DNN likely because the dataset used in Experiment 2 as true labels was annotated using the automatic annotation tools of Julius GMM.

These results show that the proposed Prosodic DAA is a more effective machine learning method for estimating the latent double articulation structure of time series data, including prosody.

\section{Experiment 3: Continuous Japanese Speech Signals following Zipf's law}\label{sec:Experiment 3}
In the third experiment, we evaluated the Prosodic DAA using continuous, more naturalistic Japanese speech signals than those in Experiment 2 from the viewpoint of distributional properties. It is widely known that words are distributed following Zipf’s law, which is a power law, in other words, in documents and utterances~\cite{zipf2016human}. 
Zipf's law is an empirical law found in many types of social, physical, and other scientific domains. In data satisfying Zipf's law, the rank--frequency distribution has an inverse relation. 
The frequency of a word is inversely proportional to its rank in the frequency table of words.
\begin{eqnarray}
 P(\geq x)\propto x^{-\alpha}
\end{eqnarray}
where $\alpha$ is a positive constant. In natural language, the word rank--frequency distribution follows $\alpha = 1$ in many cases~\cite{sano2012zipf}. 

\begin{figure}[!t]
  \begin{center}
    \includegraphics[width=\linewidth]{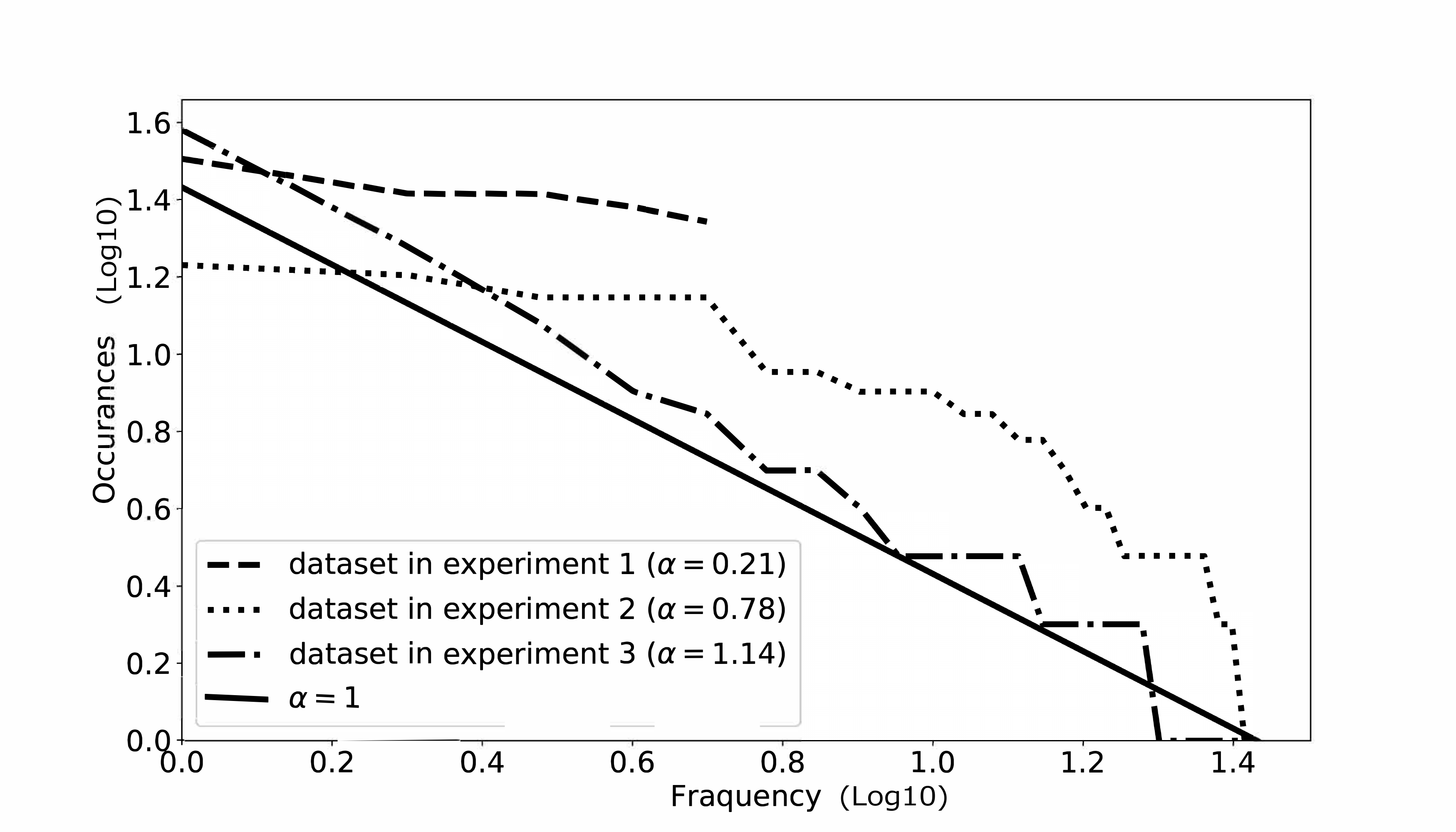}
    \caption{Log-log plot of the rank—frequency distribution of the datasets used in the experiments. The base of the logarithm for each axis is $10$.}
    \label{fig:exp3data}

  \end{center}
\end{figure}

The dataset used in Experiments 1 and 2 did not follow  Zipf’s law. Figure \ref{fig:exp3data} shows the log—log plot of the rank—frequency distributions of datasets for Experiments 1 and 2. This shows that the datasets did not follow Zipf’s law. Mathematically, the corpus following Zipf’s law has more words that appear less frequently and the distributional cues for word segmentation are more difficult to capture. Therefore, we hypothesize that prosodic cues contribute significantly to word and phoneme discovery.

\subsection{Conditions}

\begin{table*}[ht]
\begin{center}
\caption{ARI for estimated phonemes and words in Experiment 3}
\label{tab:experiment3}
\begin{tabular}{lcccc}\hline
\textbf{Method} & \textbf{Phoneme ARI} & \textbf{Word ARI} & \textbf{Trained Acoustic Model} & \textbf{True Word Dictionary} \\ \hline
Prosodic DAA (fundamental frequency $\rm F_0$ + silent pause) &0.291$\pm$0.016 & \underline{\bf0.539$\pm$0.039} \\
Prosodic DAA (silent pause) & 0.292$\pm$0.019 & \bf0.478$\pm$0.088 \\
Prosodic DAA (fundamental frequency $\rm F_0$) & \underline{\bf0.296$\pm$0.015} & 0.373$\pm$0.039 \\
NPB-DAA & \bf0.292$\pm$0.015 & 0.303$\pm$0.040 \\ \hline
Julius GMM & 0.630 & 0.822 & \checkmark \\
Julius DNN & 0.437 & 0.533 & \checkmark \\ \hline
Julius GMM with true word dictionary & 0.652 & 0.930 & \checkmark & \checkmark \\
Julius DNN with true word dictionary & 0.471 & 0.829 & \checkmark & \checkmark \\ \hline
\end{tabular} 
\end{center}
\end{table*}

The dataset used in Experiment 3 was the same type of dataset used in Experiment 2, except for word frequency distribution. The word frequency is adjusted to follow Zipf’s law\footnote{Japanese native speech dataset following Zipf’s law
 :\\\indent \url{https://github.com/EmergentSystemLabStudent/object_teaching_dataset/}}. Figure~\ref{fig:exp3data} shows the log--log plot of the rank--frequency distribution of the dataset used in Experiment 3. The figure shows that the dataset follows Zipf's law, where $\alpha =1$.
A native Japanese male speaker read aloud 42 sentences at a natural speed; the utterances were recorded. 
The sentences are about teaching the names and features of objects in the same way as those in Experiment 2.
The sentences consist of 26 Japanese phonemes and 27 words. 

The other experimental settings were same as those in Experiment 2.

\subsection{Results}
Table~\ref{tab:experiment3} shows ARIs of discovered phonemes and words. Each shows an average of over 20 trials. 
The baseline methods are the same as those in Experiment 2.

The experimental results show that the Prosodic DAA improved the word ARI compared to NPB-DAA in every condition significantly at $p = 0.05$ with t-test ($p=7.2\times 10^{-21}$, $2.1\times 10^{-06}$, and $1.3 \times 10^{-8}$ for Prosodic DAA with both features, $\rm F_0$, and silent pause, respectively ). 
In contrast, significant differences about phoneme ARI in every condition were not found.

This result shows that the prosodic cue contributes to the word segmentation task even when the target data follow Zipf's law. Comparing Tables 2 and 3, we find that the performance of NPB-DAA, which only uses distributional cues, deteriorated. 
However, we can also find that the introduction of prosodic cues in Experiment 3 improved the ARIs more than those in Experiment 2.
This suggests that when prosodic cues contribute to word discovery in natural speech signals, the distributional cues are statistically hard to capture.

\section{Conclusion} \label{sec:Conclusion}
In this study, we proposed the Prosodic DAA for discovering words directly from continuous human speech signals using statistical information and prosodic information in an unsupervised manner. For this purpose, we proposed a probabilistic generative model called the Prosodic HDP-HLM by extending the HDP-HLM. Based on the generative model, we derived an inference procedure by expanding the blocked Gibbs sampler proposed for HDP-HLM. To evaluate the performance of the proposed method, we conducted three experiments. In the first experiment, we applied the proposed method to actual human Japanese vowel speech signals. In the second experiment, the proposed method was applied to actual human Japanese utterances. In the third experiment, the proposed method was applied to Japanese utterances whose word distribution follows Zipf's law. The results showed that the proposed method could make use of prosody information and outperformed NPB-DAA in word segmentation performance. However, the phoneme discovery performance did not improve. This suggests that prosodic cues, i.e., second derivatives of $\rm F_0$ and silent pauses, do not contribute to phoneme discovery. In addition, the third experiment suggests that prosodic cues contribute to word segmentation if the distributional cues are difficult to capture, for example, the word distribution follows Zipf's law. 

Word and phoneme discovery from more natural speech signals will be a crucial challenge in our future work. We performed word and phoneme discovery from speech signals. However, we limited the number of words and phonemes in the experiment. Therefore, we did not test our method on the open-ended learning of words and phonemes. Language acquisition from speech signals, emphasizing prosodies such as infant-directed speech by human parents and natural speech signals such as daily conversation, is a topic for our future work.

Computational cost is still a problem in our method. 
Current inference procedure requires 
$O(TN^2L_{max}d^2_{max})$, where $T$ is the number of frames, $L_{max}$ is the maximum number of latent letters in a latent word, $d_{max}$ is the maximum frame length of a latent word, and $N_{max}$ is the maximum number of words~\cite{RyoOzaki2018}. 
The inference time of the dataset conditioned the maximum number of letters and words to 10 in Experiment 1 and took approximately 4 min, and the dataset conditioned the maximum number of letters and words to 50 in Experiment 2 and took approximately 30 min for Gibbs sampling 100 iterations using two Intel Xeon CPU E5-2650 v2 2.60 GHz, 8 cores, 16 threads CPUs. The inference time depends on the maximum number of words. Therefore, an improvement in the computational cost is still required for a large dataset with many words. Introducing a neural network for inference is a feasible approach to reduce the computational cost, i.e., amortized inference~\cite{kingma2013auto}. This allows us to make use of GPUs.

One direction of our future research is to develop a robot that automatically learns words and phonemes simply by speaking and responding to them. In this study, we focused on language acquisition from statistical information and prosodic information and proposed a mathematical model for language acquisition. Several studies have suggested that using co-occurrence information improves the accuracy of language acquisition~\cite{taniguchi2018unsupervised,nakamura2018ensemble}. Another direction of our future research is to combine co-occurrence cues into a double articulation analyzer and obtain a mathematical model for more accurate word and phoneme discovery.

\bibliographystyle{IEEEtran}
\bibliography{main}

\begin{IEEEbiography}[{\includegraphics[width=1in,height=1.25in,clip,keepaspectratio]{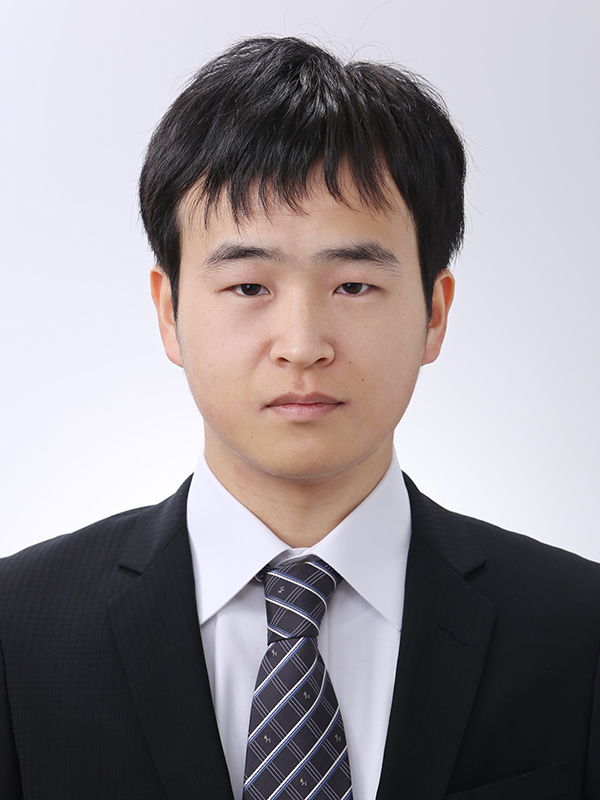}}]
{Yasuaki~Okuda} received a BE degree in Information Science and Engineering from Ritsumeikan University in 2019, and ME degree from the Graduate School
of Information Science and Engineering from Ritsumeikan University in 2021. His current research interests include machine learning and language acquisition.
\end{IEEEbiography}
\begin{IEEEbiography}[{\includegraphics[width=1in,height=1.25in,clip,keepaspectratio]{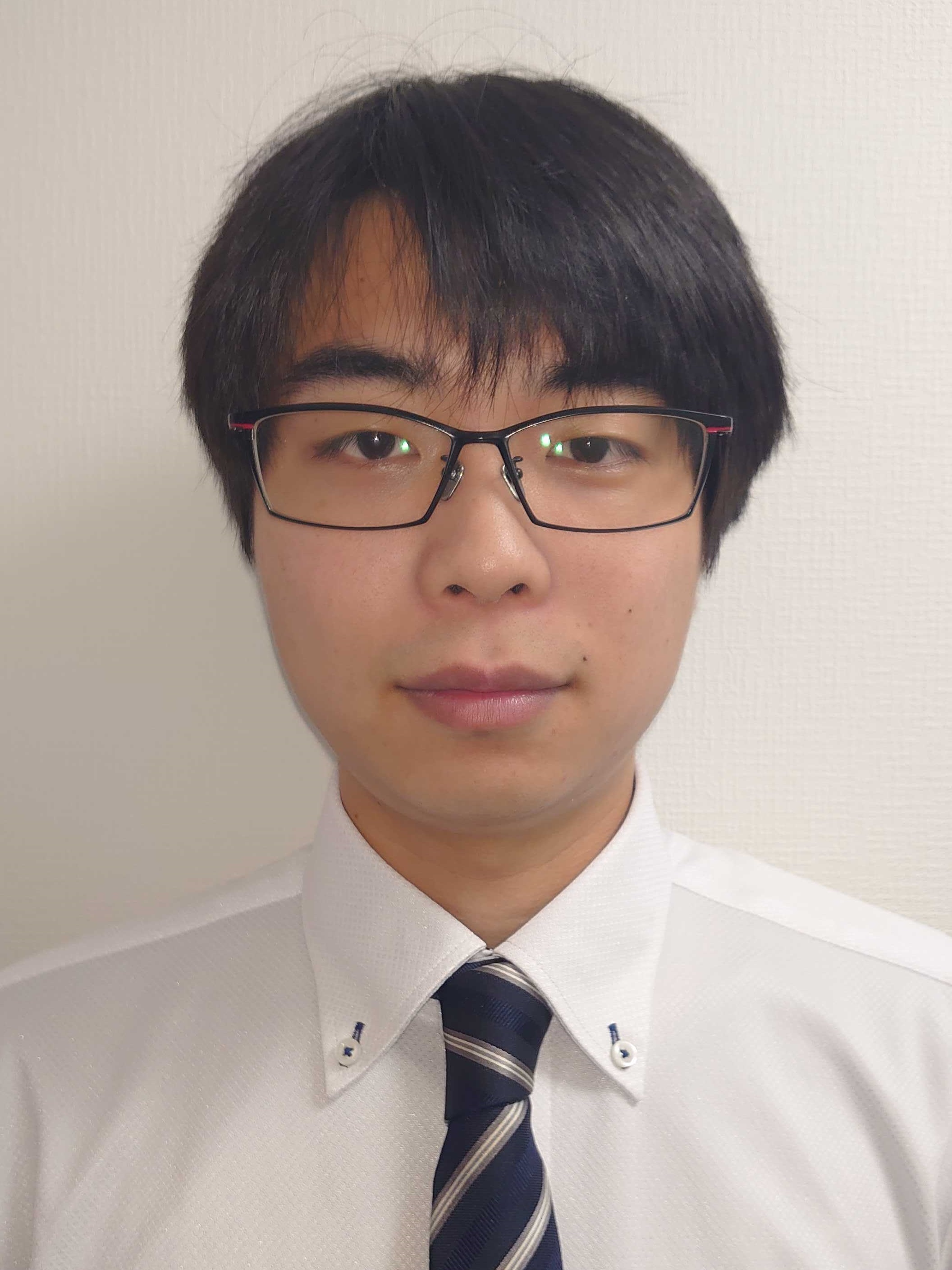}}]
{Ryo~Ozaki} received a BE degree in Information Science and Engineering from Ritsumeikan University in 2018, and ME degree from the Graduate School
of Information Science and Engineering from Ritsumeikan University in 2020. His current research interests include machine learning and language acquisition.
\end{IEEEbiography}
\begin{IEEEbiography}[{\includegraphics[width=1in,height=1.25in,clip,keepaspectratio]{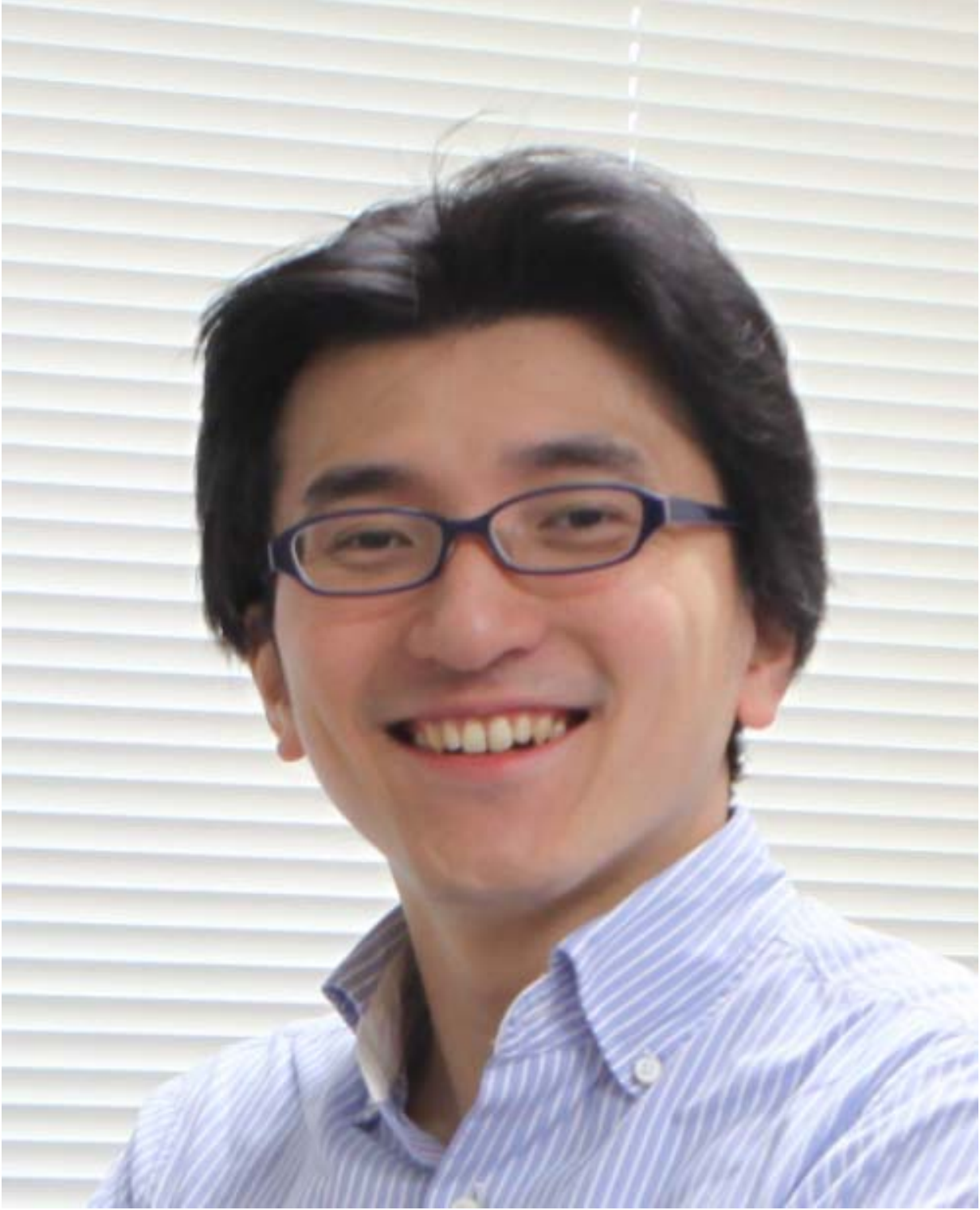}}]
{Tadahiro~Taniguchi} received his M.E. and Ph.D. degrees from Kyoto University in 2003 and 2006, respectively. He was a Japan Society for the Promotion of Science Research Fellow at the same university from 2005 to 2008. He was an Assistant Professor at the Department of Human and Computer Intelligence, Ritsumeikan University from 2008 to 2010. He was an in the same department from 2010 to 2017. He was a Visiting Associate Professor at the Department of Electrical and Electronic Engineering, Imperial College London from 2015 to 2016. Since 2017, he has been a Professor at the Department of Information and Engineering, Ritsumeikan University and a Visiting General Chief Scientist at the Technology Division of Panasonic Corporation. He has been engaged in research on machine learning, emergent systems, intelligent vehicles, and symbol emergence in robotics.
\end{IEEEbiography}
\end{document}